\documentclass{article}
\usepackage[colorlinks,linkcolor=red,citecolor=green]{hyperref}
\usepackage{spconf,amsmath,epsfig}
\usepackage{amsfonts,amssymb} 
\usepackage[ruled,linesnumbered]{algorithm2e}
\usepackage{color, soul} 
\usepackage{bbding}
\usepackage{makecell}
\usepackage{overpic}
\setulcolor{blue}

\let\OLDthebibliography\thebibliography
\renewcommand\thebibliography[1]{
  \OLDthebibliography{#1}
  \setlength{\parskip}{0pt}
  \setlength{\itemsep}{0pt plus 0.3ex}
}

\begin{document}\sloppy

\def\x{{\mathbf x}}
\def\L{{\cal L}}

\title{Multimodal Semantic-Aware Automatic Colorization with Diffusion Prior} 

%

\name{Han Wang, Xinning Chai, Yiwen Wang, Yuhong Zhang, Rong Xie, Li Song}
\address{Shanghai Jiao Tong University\\
\\
Project page: \href{https://servuskk.github.io/ColorDiff-Image/}{https://servuskk.github.io/ColorDiff-Image/}}

\maketitle

\begin{abstract}

Colorizing grayscale images offers an engaging visual experience. Existing automatic colorization methods often fail to generate satisfactory results due to incorrect semantic colors and unsaturated colors. In this work, we propose an automatic colorization pipeline to overcome these challenges. We leverage the extraordinary generative ability of the diffusion prior to synthesize color with plausible semantics. To overcome the artifacts introduced by the diffusion prior, we apply the luminance conditional guidance. Moreover, we adopt multimodal high-level semantic priors to help the model understand the image content and deliver saturated colors. Besides, a luminance-aware decoder is designed to restore details and enhance overall visual quality. The proposed pipeline synthesizes saturated colors while maintaining plausible semantics. Experiments indicate that our proposed method considers both diversity and fidelity, surpassing previous methods in terms of perceptual realism and gain most human preference.

\end{abstract}
\begin{keywords}
Automatic Colorization, Diffusion Models, Luminance Guidance, High-level Semantics
\end{keywords}
\begin{figure*}
\centering
\includegraphics[width=\textwidth]{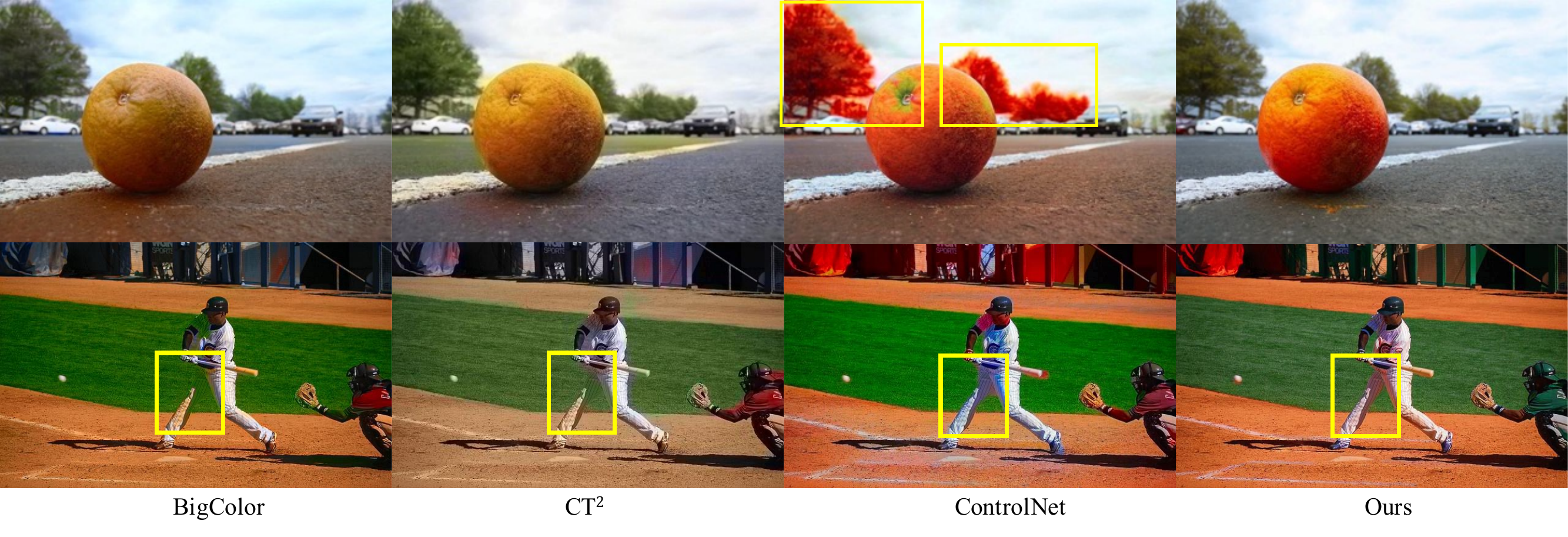}
\vspace{-0.8cm}
\caption{We achieve saturated and semantic plausible colorization for grayscale images surpassing the GAN-based(BigColor~\cite{kim2022bigcolor}), transformer-based(CT$^2$~\cite{CT2}) and diffusion-based(ControlNet~\cite{zhang2023adding}) methods.}
\label{fig:page1}
\end{figure*}
\section{Introduction}
\label{sec:intro}
\vspace{-0.1cm}
Automatic colorization synthesizes a colorful and semantically plausible image given a grayscale image. It is a classical computer vision task that has been studied for decades. However, existing automatic colorization methods cannot provide satisfactory solution due to the two main challenges: incorrect semantic colors and unsaturated colors. 

Aiming to synthesize semantically coherent and perceptually plausible colors, generative models have been extensively incorporated into relevant research. Generative adversarial networks (GAN) based~\cite{antic_deoldify_nodate, vitoria2020chromagan,kim2022bigcolor} and autoregressive-based~\cite{kumar2021colorization,CT2,huang2022unicolor} methods have made notable progress. Although the issue of incorrect semantic colors has been partially addressed, significant challenges still remain. See the yellow boxes in Figure \ref{fig:page1}, the semantic errors significantly undermine the visual quality. Recently, Denoising Diffusion Probabilistic Models(DDPM)~\cite{ho2020denoising} has demonstrated remarkable performance in the realm of image generation. With its exceptional generation capabilities, superior level of detail, and extensive range of variations, DDPM has emerged as a compelling alternative to the GAN. Moreover, the controllable generation algorithms based on the diffusion model have achieved impressive performance in various downstream tasks such as T2I~\cite{chefer2023attend}, image editing~\cite{brooks2023instructpix2pix}, super resolution~\cite{luo2023controlling}, etc. In this work, we leverage the powerful diffusion prior to synthesize plausible images that align with real-world common sense. Unfortunately, applying pre-trained diffusion models directly to this pixel-wise conditional task lead to inconsistencies~\cite{yang2023pixel} that do not accurately align with the original grayscale input. Therefore, it becomes imperative to provide more effective condition guidance in order to ensure coherence and fidelity. We align the luminance channel both in the latent and pixel spaces. Specifically, our proposed image-to-image pipeline is fine-tuned based on pre-trained stable diffusion. The pixel-level conditions are injected into the latent space to assist the denoising U-Net in producing latent codes that are more faithful to grayscale images. A luminance-aware decoder is applied to mitigate pixel space distortion. 

In addition to incorrect semantics, another challenge in this task is unsaturated colors. For example, the oranges in the first two columns of Figure \ref{fig:page1} suffer from the unsaturated colors. To moderate the unsaturated colors, priors such as categories~\cite{vitoria2020chromagan}, bounding boxes~\cite{su2020instance} and saliency maps~\cite{zhao2020scgan} have been introduced in relevant research. Based on this insight, we
adopt multimodal high-level semantic priors to help the model understand the image content and generate vivid colors. To simultaneously generate plausible semantics and  vivid colors, multimodal priors, including category, caption, and segmentation, are injected into the generation process in a comprehensive manner.

In summary, we propose an automatic colorization pipeline to address the challenges in this task. The contributions of this paper are as follows:
\vspace{-0.2cm}
\begin{itemize}
    \item We extend the stable diffusion model to automatic image colorization by introducing pixel-level grayscale conditions in the denoising diffusion. The pre-trained diffusion priors are employed to generate vivid and plausible colors.
    \vspace{-0.3cm}
    \item We design a high-level semantic injection module to enhance the model's capability to produce semantically reasonable colors.\
    \vspace{-0.3cm}
    \item A luminance-aware decoder is designed to mitigate pixel domain distortion and make the reconstruction more faithful to the grayscale input. \
    \vspace{-0.3cm}
    \item Quantitative and qualitative experiments demonstrate that our proposed colorization pipeline provides high-fidelity, color-diversified colorization for grayscale images with complex content. User study further indicates that our pipeline gain more human preference than other state-of-the-art methods.\
\end{itemize}
\vspace{-0.5cm}
\section{Related Works}
\vspace{-0.2cm}
\label{sec:related work}

Learning-based algorithms have been the mainstream of research on automatic colorization in recent years. Previous methods suffer from unsaturated colors and semantic confusion due to the lack of prior knowledge of color. In order to generate plausible colors, generative models have been applied to automatic colorization tasks, including adversarial generative networks~\cite{antic_deoldify_nodate, vitoria2020chromagan,kim2022bigcolor} and transformers~\cite{kumar2021colorization,CT2,huang2022unicolor}. Besides, ~\cite{dhariwal2021diffusion} shows that diffusion models are more creative than GAN. DDPM has achieved amazing results in diverse natural image generation. Research based on DDPM has confirmed its ability to handle a variety of downstream tasks, including colorization~\cite{wang2022zero}. To alleviate semantic confusion and synthesize more satisfactory results, priors are introduced into related research, including categories~\cite{vitoria2020chromagan}, saliency maps~\cite{zhao2020scgan}, bounding boxes~\cite{su2020instance}, etc.  \\
\vspace{-0.2cm}
\section{Method}
\vspace{-0.2cm}
\label{sec:method}
\begin{figure*}[t]
\centering
\includegraphics[width=\textwidth]{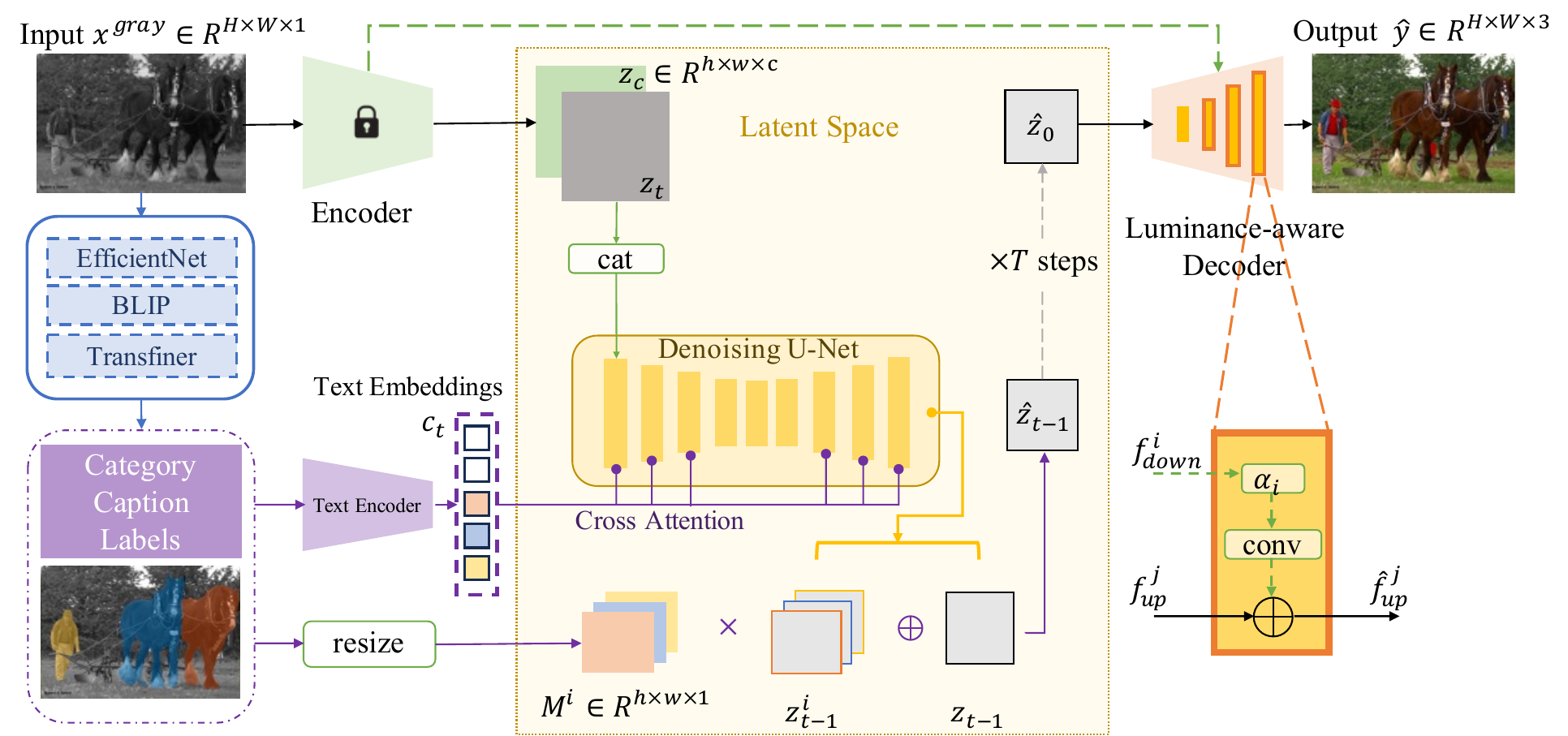}
\caption{Overview of the proposed automatic colorization pipeline. It combines a semantic prior generator (blue box), a high-level semantic guided diffusion model(yellow box), and a luminance-aware decoder (orange box).}
\label{fig:overview}
\end{figure*}
\subsection{Overview}
\vspace{-0.2cm}
A color image $y^{lab}$, represented in CIELAB color space, contains three channels: lightness channel $l$ and chromatic channels $a$ and $b$. The automatic colorization aims to recover the chromatic channels from the grayscale image: $x^{gray}\rightarrow \hat{y}^{lab}$.

In this work, we propose an automatic colorization pipeline for natural images based on stable diffusion. The pipeline consists of two parts: a variational autoencoder~\cite{kingma2013auto} and a denoising U-Net. Explicitly, the VAE is for the transformation between pixel space $x\in \mathcal{R}^{H\times W\times 3}$ and latent space $z\in \mathcal{R}^{h\times w\times c}$. While the denoising U-Net applies DDPM in the latent space to generate an image from Gaussian noise. 

The framework of our pipeline is shown in Figure~\ref{fig:overview}. First, the VAE encodes grayscale image $x^{gray}$ into latent code $z_c$. Next, the T-step diffusion process generates a clean latent code $z_0$ from Gaussian noise  $z_T$  under the guidance of image latent $z_c$ and high-level semantics. Finally, $z_0$ is reconstructed by a luminance-aware decoder to obtain the color image $\hat{y}$. 

The pixel-level grayscale condition and high-level semantic condition for denoising process are introduced in the latent space, shown in the yellow box in Figure \ref{fig:overview}. We elaborate on the detailed injections of these conditions in Section \ref{sec: gray} and Section \ref{sec: semantic}, respectively. As for the reconstruction processes, the detailed designs of the luminance-aware decoder are described in Section \ref{sec: decoder}.

\begin{figure*}[t]
\centering
\includegraphics[width=\textwidth]{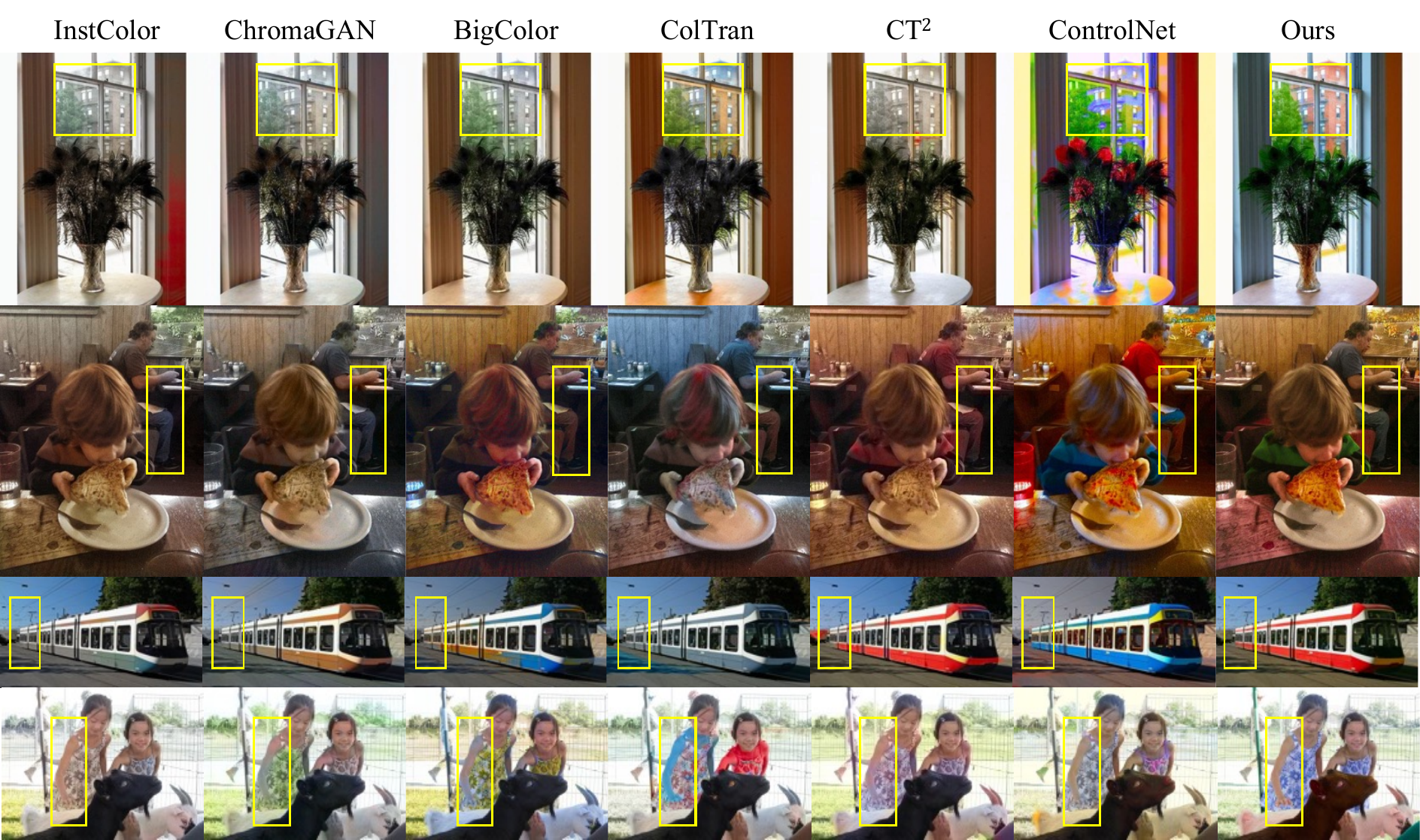}
\vspace{-20pt}
\caption{Qualitative comparisons among InstColor~\cite{su2020instance}, ChromaGAN~\cite{vitoria2020chromagan}, BigColor~\cite{kim2022bigcolor}, ColTran~\cite{kumar2021colorization}, CT$^2$~\cite{CT2}, ControlNet~\cite{zhang2023adding} and Ours. More results are provided on 
\href{https://servuskk.github.io/ColorDiff-Image/}{https://servuskk.github.io/ColorDiff-Image/}.}
\label{fig:base}
\vspace{-0.4cm}
\end{figure*}
\vspace{-0.2cm}
\subsection{Colorization Diffusion Model}\label{sec: gray}
Large-scale diffusion model has the capability to generate high-resolution images with complex structures. While naive usage of diffusion priors generates serious artifacts, we introduce pixel-level luminance information to provide detailed guidance.

Specifically, we use encoded grayscale image $z_c$ as control condition to enhance U-Net's understanding of luminance information in the latent space. To involve the grayscale condition in the entire diffusion process, we simultaneously input the latent code $z_{t}$ generated in the previous time step and the noise-free grayscale latent code $z_c$ into the input layer of U-Net at each time step $t$:
\vspace{-0.1cm}
\begin{equation}
    z^\prime_{t} = \mathtt{conv_{1\times1}}(\mathtt{concat}(z_{t} ,z_c))
\end{equation}
In this way, we take advantage of the powerful generative capabilities of stable diffusion while preserve the grayscale condition. The loss function for our denoising U-Net is defined in a similar way to stable diffusion~\cite{rombach2022high}:
\begin{equation}
    \mathcal{L}=\mathbb{E}_{z,z_c,c,\epsilon\sim \mathcal{N}(0,1),t}[||\epsilon-\epsilon_\theta(z_t,t,z_c,c)||_2^2]
\end{equation}
where $z$ is the encoded color image, $z_c$ is the encoded grayscale image, $c$ is the category embedding, $\epsilon$ is a noise term, $t$ is the time step, $\epsilon_\theta$ is the denoising U-Net, $z_t$ is the noisy version of $z$ at time step $t$.

\subsection{High-level Semantic Guidance}\label{sec: semantic}

To alleviate semantic confusion and generate vivid colors, we design a high-level semantic guidance module for inference. As shown in Figure \ref{fig:overview}, the multimodal semantics are generated by the pre-trained semantic generator in the blue box. Afterwards, text and segmentation priors are injected into the inference process through cross attention and segmentation guidance respectively, as shown in the yellow box in Figure \ref{fig:overview}. Specifically, given the grayscale image $x^{gray}$, the semantic generator produce the corresponding categories~\cite{koonce2021efficientnet},  captions~\cite{li2022blip} and segmentations~\cite{ke_mask_2021}. The category, caption, and segmentation labels are in textual form, while the segmentation masks are binary masks.

For textual priors, the CLIP~\cite{radford2021learning} encoder is employed to generate the text embedding $c_t$. The text embedding guidance is applied in denoising U-Net via cross-attention mechanism. Given the timestep $t$, the concatenated noisy input $z_t$ and the text condition $c_t$, the latent code $z_{t-1}$ is produced by the Colorization Diffusion Model(CDM):
\begin{equation}
  z_{t-1} = CDM(z_t,t,z_c,c_t)  
\end{equation}

For segmentation priors, we use the pre-trained transfiner~\cite{ke_mask_2021} to generate paired segmentation masks $\mathcal{M}$ and labels $\mathcal{L}$. For each instance, we first resize the binary mask $\mathcal{M}^i \in \mathcal{R}^{H\times W\times 1}$ to align the latent space. The resized mask is represented as $M^i \in \mathcal{R}^{h\times w\times 1}$. Then we use the CDM to yield the corresponding latent code $z^i_{t-1}$ of the masked region:
\begin{equation}
  z^i_{t-1} = CDM(z_t,t,z_c\times M^i,L^i)  
\end{equation}
Finally, we combine the original latent code $z_{t-1}$ and the instances to yield the segment-aware latent code $\hat{z}_{t-1}$:
\begin{equation}
  \hat{z}_{t-1} = \sum_{i=1}^{i=k} [z_{t-1}\times (1-M^i) + z^i_{t-1}\times M^i]
\end{equation}

We set a coefficient $i\in [0,1]$ to control the strength of segmentation guidance. The threshold is defined as $T_{th}=T\times(1-i)$. The segmentation mask is used to guide the synthesis process at inference time step $t>T_{th}$. We set $i=0.3$ for the experiment. Users have the flexibility to select a different value based on their preferences.

\subsection{Luminance-aware Decoder}\label{sec: decoder}
As the downsampling to latent space inevitably lose detailed structures and textures, we apply the luminance condition to the reconstruction process and propose a luminance-aware decoder. To align the latent space with stable diffusion, we freeze the encoder. The intermediate grayscale features obtained in the encoder are added to the decoder through skip connections. 

Specifically, the intermediate features $f_{down}^i$ generated by the first three downsample layers of the encoder are extracted. These features are convolved, weighted, and finally added to the corresponding upsample layers of the decoder:
\begin{equation}
\hat{f}_{up}^j=f_{up}^j + \alpha_i \cdot \mathtt{conv}(f_{down}^i),  i=0,1,2; j=3,2,1
\end{equation}

We adopt L2 loss $\mathcal{L}_2$ and perceptual loss~\cite{johnson2016perceptual} $\mathcal{L}_p$ to train the  luminance-aware decoder:
\begin{equation}\label{eq:loss-d}
\vspace{-0.2cm}
    \mathcal{L}=\mathcal{L}_2 + \lambda_p\mathcal{L}_p
\vspace{-0.1cm}
\end{equation}
\vspace{-0.3cm}
\section{Experiment}
\vspace{-0.2cm}
\subsection{Implementation}
We train the denoising U-Net and luminance-aware decoder separately. Firstly, we train the denoising U-Net on the imagenet~\cite{DenDon09Imagenet} training set at the resolution of $512\times 512$. We initialize the U-Net using the pre-trained weights of ~\cite{rombach2022high}. The learning rate is fixed at $5e-5$. We use the classifier-free guidance~\cite{ho2022classifierfree} strategy and set the conditioning dropout probability to $0.05$. The model is updated for 20K iterations with a batch size of 16. Then we train the luminance-aware decoder on the same dataset and at the same resolution. The VAE is initialized using the pre-trained weights of ~\cite{rombach2022high}. We fix the learning rate at $1e-4$ for 22,500 steps with a batch size of 1. We set the parameter $\lambda_p$ in Eq.(\ref{eq:loss-d}) to $0.1$. Our tests are conducted on the COCO-Stuff~\cite{cocodataset} val set containing 5,000 images of complex scenes. At inference, we adopt DDIM sampler~\cite{song2020denoising} and set the inference time step $T=50$. We conduct all experiments on a single Nvidia GeForce RTX 3090 GPU.
\vspace{-0.2cm}
\subsection{Comparisons}\label{sec:comp}

We compare with 6 state-of-the-art automatic colorization methods including 3 types: 1) GAN-based method: InstColor~\cite{su2020instance}, ChromaGAN~\cite{vitoria2020chromagan}, BigColor~\cite{kim2022bigcolor}, 2)Transformer-based method: ColTran~\cite{kumar2021colorization}, CT$^2$~\cite{CT2}, 3) Diffusion-based method: ControlNet~\cite{zhang2023adding}.\\
\textbf{Qualitative Comparison.}
We show visual comparison results in Figure \ref{fig:base}. The images in the first and second rows indicate the ability of the models to synthesise vivid colors. Both GAN-based and transformer-based algorithms suffer from unsaturated colors. Although ControlNet synthesises saturated colors, the marked areas contain significant artifacts. Images in the third and forth rows demonstrate the ability of the models to synthesise semantically reasonable colors. InstColor, ChromaGAN, BigColor, CT$^2$ and ControlNet fail to maintain the color continuity of the same object(discontinuity of colors between the head and tail of the train, hands and shoulders of the girl). While ColTran yields colors that defy common sense (blue shadows and blue hands). In summary, our method provides vivid and semantically reasonable colorization results.\\
\begin{table}[t]
\begin{center}
\vspace{-0.6cm}
\caption{Quantitative comparison results.} \label{tab:Quantitative}
\begin{tabular}{|c|ccc|}
  \hline
  Method & FID$\downarrow$ & Colorful$\uparrow$& PSNR$\uparrow$ \\
  \hline
  InstColor~\cite{su2020instance} & $14.40$ & $27.00$& $\textbf{23.85}$  \\
ChromaGAN~\cite{vitoria2020chromagan} & $27.46$ &$27.06$ & $23.20$ \\
BigColor~\cite{kim2022bigcolor} & $10.24$& $39.65$& $20.86$  \\
ColTran~\cite{kumar2021colorization} & $15.06$ & $34.31$& $22.02$  \\
CT$^2$~\cite{CT2} & $25.87$ & $39.64$& $22.80$  \\
ControlNet~\cite{zhang2023adding}& $10.86$ & $\textbf{45.09}$& $19.95$  \\
Ours & $\textbf{9.799}$& $41.54$& $21.02$ \\
  \hline
\end{tabular}
  \vspace{-0.2cm}
\end{center}
\vspace{-0.2cm}
\end{table}
\vspace{-0.2cm}
\begin{figure}[t]
\centering
\includegraphics[width=0.5\textwidth]{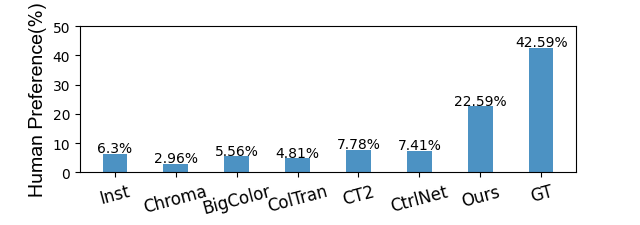}
\vspace{-0.8cm}
\caption{User evaluations.}
\label{fig:us}
\vspace{-0.5cm}
\end{figure}
\begin{table}[t]
\begin{center}
\vspace{-0.6cm}
\caption{Quantitative comparison of ablation studies.} \label{tab:ablation}
\begin{tabular}{|c|cc|cc|}
  \hline
  Exp. &\makecell[c]{Luminance-\\aware decoder}&\makecell[c]{High-level\\guidance}&FID$\downarrow$ & Colorful$\uparrow$\\
  \hline
  $(a)$ &\Checkmark & & $10.05$ & $33.73$ \\
 $(b)$ &&\Checkmark & $9.917$ & $42.55$ \\
Ours &\Checkmark &\Checkmark & $9.799$& $41.54$\\
  \hline
\end{tabular}
\end{center}
\vspace{-0.5cm}
\end{table}

\noindent\textbf{User Study.} To reflect human preferences, we randomly select 15 images from the COCO-Stuff val set for user study. For each image, the 7 results and \textbf{ground truth} are displayed to the user in a random order. We asked 18 participants to choose their top three favorites. Figure \ref{fig:us} shows the proportion of the Top 1 selected by users. Our method has a vote rate of 22.59\%, which significantly outperforms other methods.

\begin{figure}[t]
\centering
\begin{minipage}[b]{.49\linewidth}
  \centering
  \includegraphics[width=\textwidth]{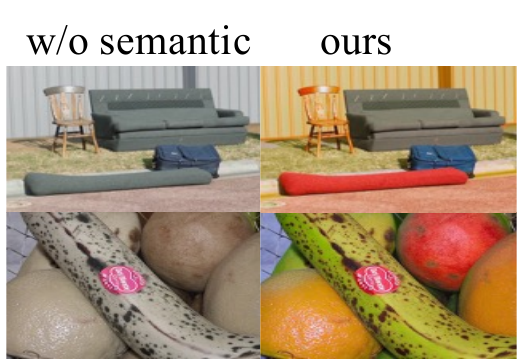}
  \centerline{(a)High-level guidance.}\medskip
\end{minipage}
\begin{minipage}[b]{.49\linewidth}
  \centering
  \includegraphics[width=\textwidth]{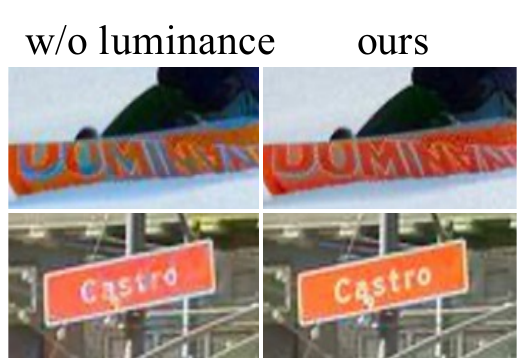}
  \centerline{(b)Luminance-aware decoder.}\medskip
\end{minipage}
\vspace{-0.5cm}
\caption{Visual comparison from ablation studies.}
\label{fig:ablation}
\vspace{-0.5cm}
\end{figure}
\noindent\textbf{Quantitative Comparison.} We use Fréchet Inception Distance (FID) and colorfulness~\cite{hasler2003measuring} to evaluate image quality and vividness. These two metrics have recently been used to evaluate the colorization algorithm~\cite{kim2022bigcolor, wu2021towards} . Considering that colorization is an ill-posed problem, the ground-truth dependent metric PSNR used in previous works does \textbf{not} accurately reflect the quality of image and color generation~\cite{kumar2021colorization, wu2021towards,xia2022disentangled}, and the comparison here is for reference. As shown in Table \ref{tab:Quantitative}, our proposed method demonstrates superior performance in terms of FID when compared to the state-of-the-art algorithms. Even though ControlNet outperforms our algorithm in the colourful metric, the results shown in the qualitative comparison indicate that the artefacts are meaningless and negatively affect the visual effect of the image.
\vspace{-0.3cm}
\subsection{Ablation Studies}

The significance of the main components of the proposed method is discussed in this section. The quantitative and visual comparisons are presented in Table \ref{tab:ablation} and Figure \ref{fig:ablation}. \\
\textbf{High-level Semantic Guidance.} We discuss the impact of high-level semantic guidance on model performance. The visuals shown in Figure \ref{fig:ablation}(a) demonstrate our high-level guidance improves saturation of synthesised colors and mitigates failures caused by semantic confusion. The quantitative scores in Table \ref{tab:ablation} confirm the significant improvement in both color vividness and perceptual quality introduced by the high-level semantic guidance.\\
\textbf{Luminance-aware Decoder.} The pipeline equipped with a luminance-aware decoder facilitates the generation of cognitively plausible colors. As shown in the first row of Figure \ref{fig:ablation}(b), the artifacts are suppressed. Furthermore, the incorporation of this decoder yields a positive impact on the retrieval of image details, as demonstrated by the successful reconstruction of textual elements in the second row of Figure \ref{fig:ablation}(b). Consequently, the full model outperforms the alternative in terms of FID. It is reported a slight decrease in colorfulness score after incorporating luminance awareness which can be attributed to the suppression of outliers, as discussed in Section \ref{sec:comp} regarding the analysis of the ControlNet.

\section{Conclusion}
\vspace{-0.3cm}
In this study, we introduce an novel automatic colorization pipeline that harmoniously combines color diversity with fidelity.  It generate plausible and saturated colors by leveraging powerful diffusion priors with the proposed luminance and high-level semantic guidance. Besides, we design a luminance-aware decoder to restore image details and improve color plausibility. Experiments demonstrate that the proposed pipeline outperforms previous methods in terms of perceptual realism and attains the highest human preference compared to other algorithms.

\section{Acknowledgement}
\vspace{-0.3cm}
This work was supported by National Key R\&D Project of China(2019YFB1802701), MoE-China Mobile Research Fund Project(MCM20180702), the Fundamental Research Funds for the Central Universities; in part by the 111 project under Grant B07022 and Sheitc No.150633; in part by the Shanghai Key Laboratory of Digital Media Processing and Transmissions.


\vspace{-0.5cm}
\bibliographystyle{IEEEbib}
\small
\bibliography{icme2023template}

\end{document}